\pdfoutput=1

\documentclass[11pt]{article}

\usepackage[]{acl2023}

\usepackage{times}
\usepackage{latexsym}

\usepackage[T1]{fontenc}

\usepackage[utf8]{inputenc}

\usepackage{microtype}

\usepackage{graphicx}
\usepackage{booktabs}
\usepackage{multirow}
\usepackage{makecell}
\usepackage{xcolor, soul}

\definecolor{light_blue}{rgb}{0.80,0.85,1.0}
\definecolor{light_red}{rgb}{1.0,0.85,0.80}

\title{Plan-and-Solve Prompting: Improving Zero-Shot Chain-of-Thought \\ Reasoning by Large Language Models
}

\author{
    ~~Lei Wang$^{1}$
   ~~Wanyu Xu$^{2}$
   ~~Yihuai Lan 
   ~~Zhiqiang Hu$^{3}$ 
   ~~Yunshi Lan$^{4}$ \\ \bf
   ~~Roy Ka-Wei Lee$^{3}$
   ~~Ee-Peng Lim$^{1}$\thanks{~~Corresponding author.}\\ 
   $^{1}$Singapore Management University \\
   $^{2}$Southwest Jiaotong University \\
   $^{3}$Singapore University of Technology and Design\\
   $^{4}$East China Normal University\\
}

\begin{document}
\maketitle

\begin{abstract}
Large language models (LLMs) have recently been shown to deliver impressive performance in various NLP tasks. 
To tackle multi-step reasoning tasks, few-shot chain-of-thought (CoT) prompting 
includes a few manually crafted step-by-step reasoning demonstrations which 
enable LLMs to explicitly generate reasoning steps and improve their reasoning task accuracy. 
To eliminate the manual effort, Zero-shot-CoT concatenates the target problem statement with ``\textit{Let's think step by step}'' as an input prompt to LLMs.
Despite the success of Zero-shot-CoT, it still suffers from three pitfalls: calculation errors,  missing-step errors, and semantic misunderstanding errors.
To address the missing-step errors, we propose Plan-and-Solve (PS) Prompting.
It consists of two components: first, devising a plan to divide the entire task into smaller subtasks, and then carrying out the subtasks according to the plan.
To address the calculation errors and improve the quality of generated reasoning steps, we extend PS prompting with more detailed instructions and derive PS+ prompting.
We evaluate our proposed prompting strategy on ten datasets across three reasoning problems.
The experimental results over GPT-3 show that our proposed zero-shot prompting consistently outperforms Zero-shot-CoT across all datasets by a large margin, is comparable to or exceeds Zero-shot-Program-of-Thought Prompting, and has comparable performance with 8-shot CoT prompting on the math reasoning problem. The code can be found at \href{https://github.com/AGI-Edgerunners/Plan-and-Solve-Prompting}{https://github.com/AGI-Edgerunners/Plan-and-Solve-Prompting}.

\end{abstract}

\section{Introduction}
\label{sec:intro}

\begin{figure}[t] 
	\centering
	\includegraphics[scale = 0.4]{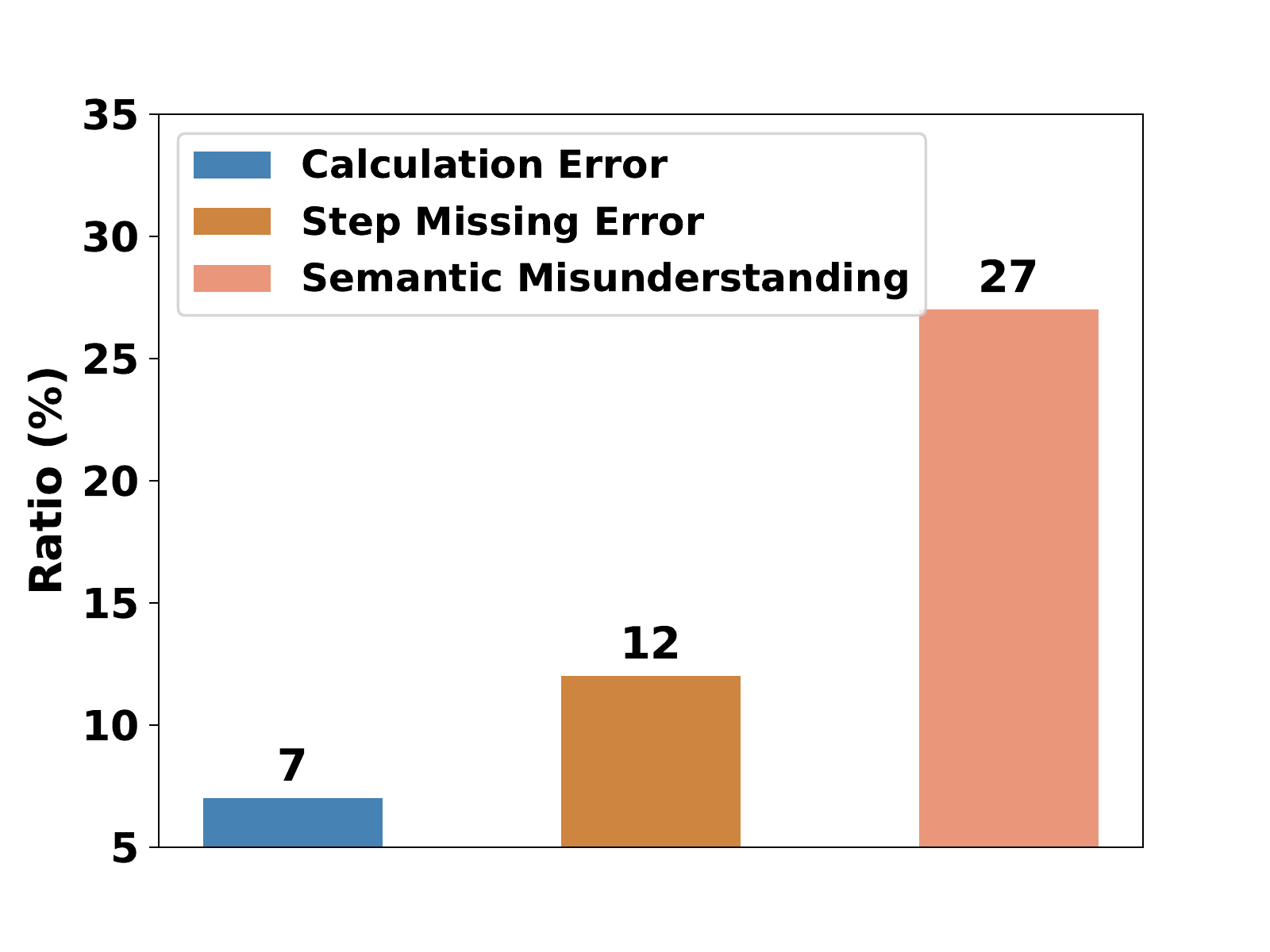} 
	\vspace{-15pt}
	\caption{Error analysis of 46 GSM8K problems with incorrect answers returned by Zero-shot-CoT using GPT-3 LLM. Following \citet{cot_wei} and \citet{wang2022towards}, we assign  ``Calculation Error'' (7\%), ``Step Missing Error'' (12\%), or ``Semantic misunderstanding Error'' (27\%) to each incorrect answer.}
	\label{fig:intro_bar}
	\vspace{-10pt}
\end{figure}

Large language models (LLMs)~\cite{brown2020language, thoppilan2022lamda, palm} have recently proven highly effective in various NLP tasks.
Unlike the previous pre-trained language models (PTMs)~\cite{bert, liu2019roberta}, these LLMs are typically provided as a service, with no access to model parameters due to commercial considerations and potential risks of misuse~\cite{sun2022blackbox}.
Thus, it is challenging to fine-tune LLMs for downstream tasks~\citep{he2021towards_finetuning, houlsby2019parameter_finetuning, bert}.
Instead, we leverage LLMs to solve complex reasoning problems by eliciting their strong reasoning abilities over their embedded knowledge using instructions (or trigger sentences).  So far, LLMs have shown impressive abilities to solve new reasoning problems by simply conditioning them on a few illustrative examples (i.e., few-shot learning) or a prompt to solve new problems without illustrative examples (i.e., zero-shot learning). 

To tackle multi-step complex reasoning tasks using LLMs, \citet{cot_wei} proposes few-shot chain-of-thought (CoT) prompting, 
which enables LLMs to explicitly generate the intermediate reasoning steps before predicting the final answer with a few manual step-by-step reasoning demonstration examples.
In \citep{kojima2022large}, Zero-shot CoT eliminates the need for manually crafted examples in prompts by appending ``\textit{Let's think step by step}'' to the target problem fed to LLMs such as GPT-3. This simple prompting strategy surprisingly enables LLMs to yield performance similar to few-shot CoT prompting.

\begin{figure*}[t]
  \centering
  \includegraphics[width=1.0\linewidth]{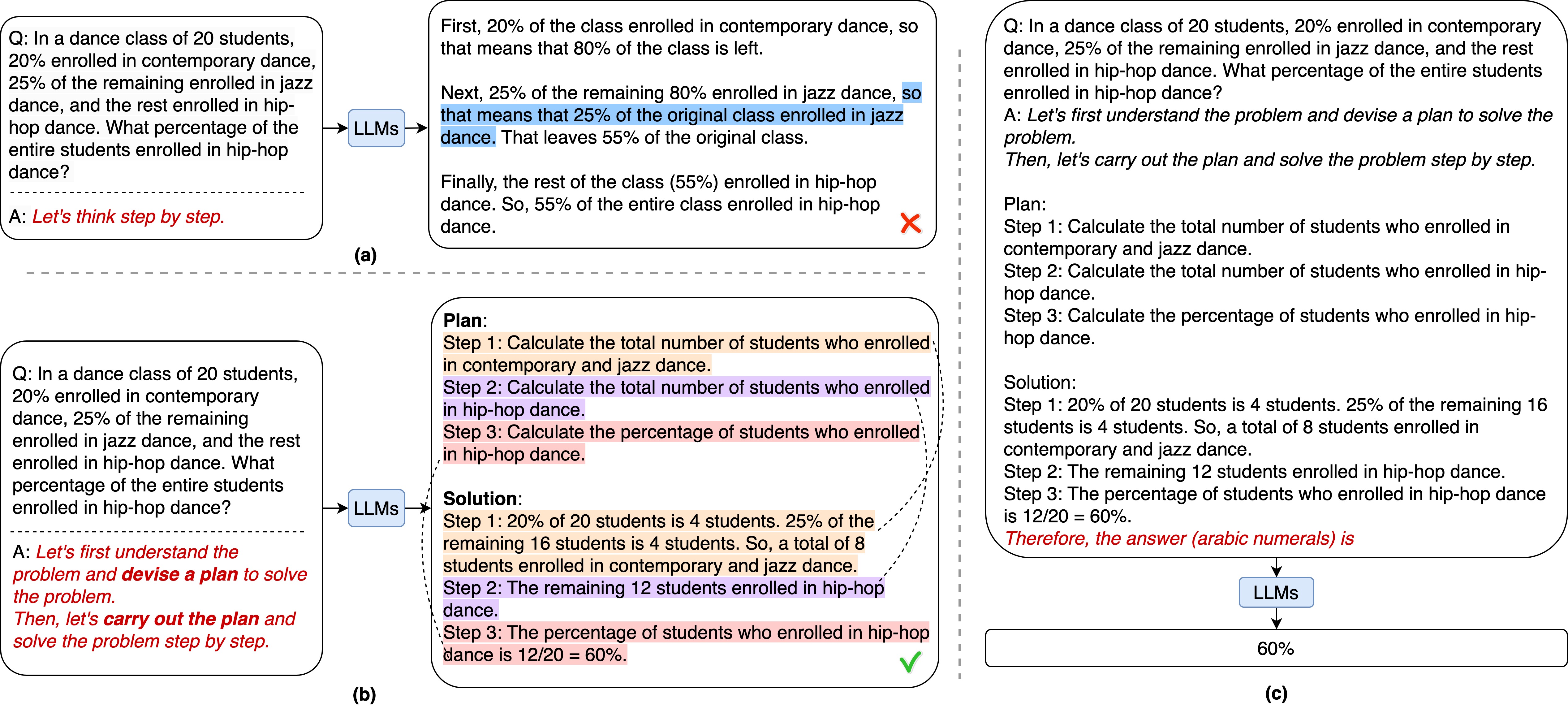}
  \vspace{-20pt}
  \caption{
    Example inputs and outputs of GPT-3 with (a) Zero-shot-CoT prompting, (b) Plan-and-Solve (PS) prompting, and (c) answer extraction prompting. While Zero-shot-CoT encourages LLMs to generate multi-step reasoning with ``\textit{Let's think step by step}'', it may still generate wrong reasoning steps when the problem is complex. Unlike Zero-shot-CoT, PS prompting first asks LLMs to devise a plan to solve the problem by generating a step-by-step plan and carrying out the plan to find the answer.
  }
  \label{fig_1}
\vspace{-10pt}
\end{figure*}

Despite the remarkable success of Zero-shot-CoT in solving multi-step reasoning tasks, its results on a sample of $100$ arithmetic test examples still point to three pitfalls (as shown in Figure~\ref{fig:intro_bar}): 
(i) Calculation errors (in $7\%$ of test examples): These are errors in the calculation leading to wrong answers;
(ii) Missing Step errors (in $12\%$ of test examples):  These occur when some intermediate reasoning step(s) is missed-out especially when there are many steps involved;
(iii) Semantic misunderstanding (in $27\%$ of test examples):  There are other errors in semantic understanding of the problem and coherence of reasoning steps likely to be caused by the insufficient capability of LLMs. 

To address the issue of Zero-shot-CoT caused by missing reasoning steps, we propose Plan-and-Solve (PS) Prompting.
It consists of two components: first, devising a plan to divide the entire task into smaller subtasks, and then carrying out the subtasks according to the plan.
In our experiments, we simply replace ``\textit{Let's think step by step}'' of Zero-shot-CoT with ``\textit{Let’s first understand the problem and devise a plan to solve the problem. Then, let’s carry out the plan and solve the problem step by step}'' (see Figure~\ref{fig_1} (b)). 

\begin{figure*}[t]
  \centering
  \includegraphics[width=1.0\linewidth]{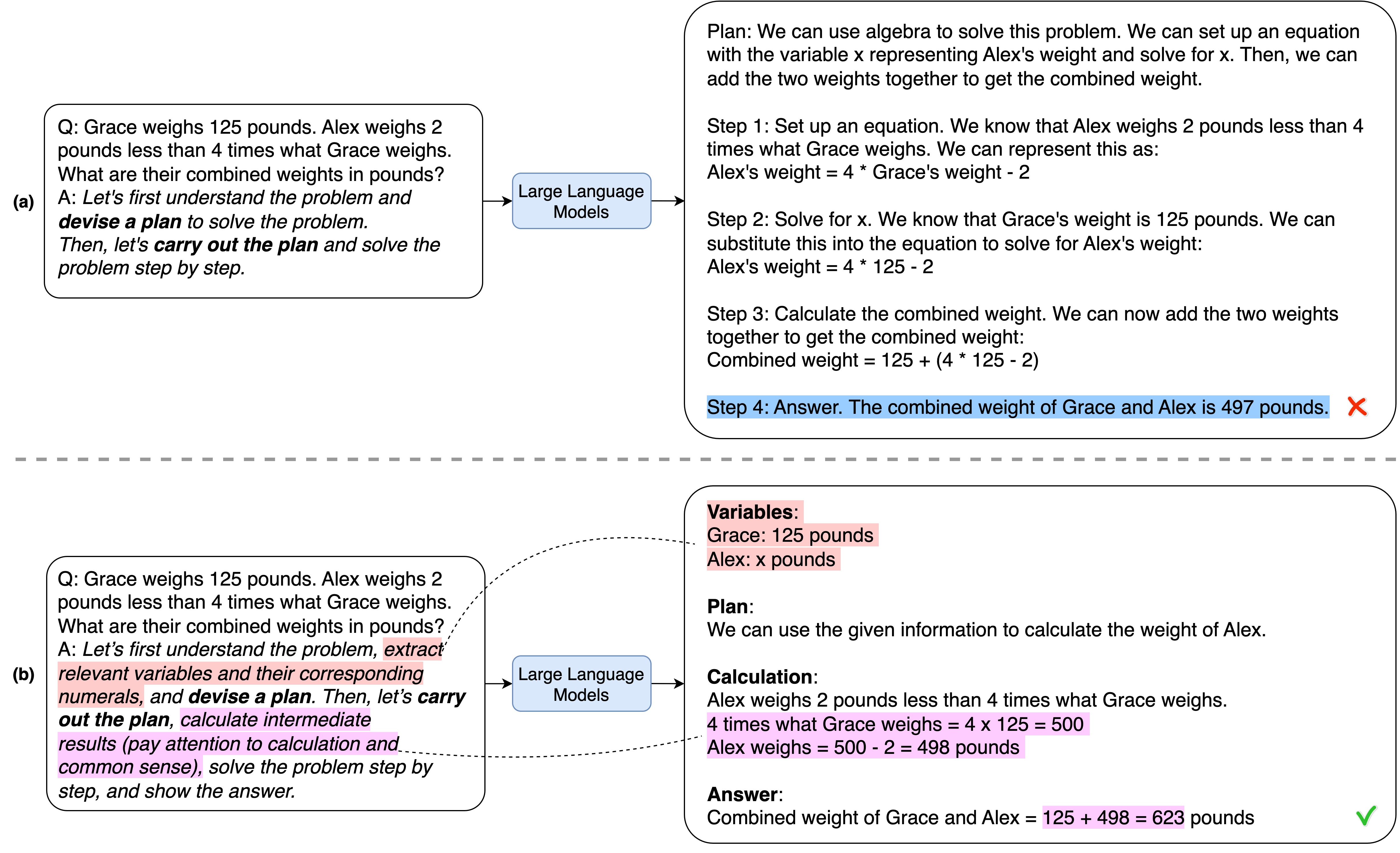}
  \vspace{-20pt}
  \caption{
    Example inputs and outputs of GPT-3 with (a) Plan-and-Solve (PS) Prompting and (b) Plan-and-Solve prompting with more detailed instructions (PS+ prompting). PS+ prompting greatly improves the quality of the generated reasoning process.
  }
  \label{fig_2}
\vspace{-10pt}
\end{figure*}

To address the calculation errors of Zero-shot-CoT and improve the quality of generated reasoning steps, we add more detailed instructions to PS prompting. Specifically, we extend it with ``\textit{extract relevant variables and their corresponding numerals}'' and ``\textit{calculate intermediate results (pay attention to calculation and commonsense)}'' instructions. This prompting variant is called the PS+ prompting strategy (see Figure~\ref{fig_2} (b)).
Despite its simplicity, PS+ strategy greatly improves the quality of the generated reasoning process. 
Moreover, this prompting strategy can be easily customized to solve a variety of problems other than math reasoning, such as commonsense and symbolic reasoning problems.

We evaluate our proposed prompting on six math reasoning datasets, including AQuA~\cite{aqua}, GSM8K~\cite{gsm8k}, MultiArith, AddSub, SingleEq, and SVAMP~\cite{svamp}, two commonsense reasoning datasets (CommonsenseQA~\cite{commonsenseqa} and StrategyQA~\cite{strategyqa}), and two symbolic reasoning datasets (Last Letter and Coin Flip~\cite{cot_wei}).
The results of our experiments with GPT-3 show that our proposed Zero-shot-PS+ prompting consistently outperforms Zero-shot-CoT across all reasoning problems and datasets by a large margin, and is comparable to or exceeds Zero-shot-Program-of-Thought (PoT) Prompting~\cite{chen2022program}).
Furthermore, although PS+ prompting does not require manual demonstration examples, it has a performance similar to an 8-shot CoT prompting in arithmetic reasoning.

Overall, our results suggest that (a) Zero-shot PS prompting is capable of generating a higher-quality reasoning process than Zero-shot-CoT prompting, as the PS prompts provide more detailed instructions guiding the LLMs to perform correct reasoning tasks;
(b) Zero-shot PS+ prompting outperforms Few-shot manual-CoT prompting on some datasets, indicating that in some instances it has the potential to outperform manual Few-shot CoT prompting, which hopefully will spark further development of new CoT prompting approaches to elicit reasoning in LLMs.

\section{Plan-and-Solve Prompting}
\label{sec:proposal}

\paragraph{Overview.} We introduce PS prompting, a new zero-shot CoT prompting method, which enables LLMs to explicitly devise a plan for solving a given problem and generate the intermediate reasoning process before predicting the final answer for the input problem.
As opposed to prior few-shot CoT approaches where step-by-step few-shot demonstration examples are included in the prompt, the zero-shot PS prompting method does not require demonstration examples,  and its prompt covers the problem itself and a simple trigger sentence.
Similar to Zero-shot-CoT, Zero-shot PS prompting consists of two steps.  In step 1, the prompt first makes an inference using the proposed prompting template to generate the reasoning process and the answer to a problem. In step 2, it extracts the answer for evaluation by using the answer extraction prompting, such as ``Therefore, the answer (arabic numerals) is''. 

\subsection{Step 1: Prompting for Reasoning Generation}
To solve the input problem while avoiding errors resulting from incorrect calculation and missing reasoning steps, this step aims to construct templates to meet the following two criteria:
\begin{itemize}
    \item The templates should elicit LLMs to determine subtasks and accomplish the subtasks.
    \item The templates should guide LLMs to pay more attention to calculations and intermediate results and to ensure that they are correctly performed as much as possible.
\end{itemize}

To meet the first criterion, we follow Zero-shot-CoT and first convert the input data example into a prompt with a simple template ``Q: \texttt{[X]}. A: \texttt{[T]}''. 
Specifically, the input slot \texttt{[X]} contains the input problem statement and a hand-crafted instruction is specified in the input slot \texttt{[T]} to trigger LLMs to generate a reasoning process that includes a plan and steps to complete the plan.  In Zero-shot-CoT, the instruction in the input slot \texttt{[T]} includes the trigger instruction `\textit{Let's think step by step}''.  Our Zero-shot PS prompting method instead includes the instructions ``\textit{devise a plan}'' and ``\textit{carry out the plan}'' as shown in Figure 2(b).  Thus, the prompt would be ``Q: \texttt{[X]}. A: \textit{\underline{Let’s first understand the problem} and \underline{devise a plan to solve the problem}. \underline{Then, let’s} \underline{carry out the plan and solve the problem step by} \underline{step}}.''

We then pass the above prompt to the LLM which subsequently outputs a reasoning process. In accordance with Zero-shot-CoT, our method uses the greedy decoding  strategy (1 output chain) for generating output by default. 

To meet the second criterion, we extend the plan-based trigger sentence with more detailed instructions.
Specifically, ``\textit{pay attention to calculation}'' is added to the trigger sentence to request the LLMs to perform calculations as accurately as possible.
To reduce errors resulting from missing necessary reasoning steps, we include ``\textit{extract relevant variables and their corresponding numerals}'' to explicitly instruct the LLMs not to ignore relevant information in the input problem statement.
We hypothesize that if the LLMs leave out the relevant and important variables, it is more likely to miss out relevant reasoning steps.
Correlation analysis of generated content of variable and the missing reasoning step errors,  shown in Figure~\ref{fig:correlation}, empirically supports this hypothesis (correlation value is less than 0). 
Additionally, we add ``\textit{calculate intermediate results}'' to the prompt to enhance LLM's ability to generate relevant and important reasoning steps.
The specific example is illustrated in Figure 3(b).
At the end of Step 1, LLM generates the reasoning text which includes the answer.  For example, the generated reasoning text in Figure 3(b) includes ``\textit{Combined weight of Grace and Alex = 125 + 498 = 623 pounds}''.
The strategy of adding specific descriptions to the trigger sentence represents a new way to improve zero-shot performance on complex reasoning.

\subsection{Step 2: Prompting for Answer Extraction}

Similar to Zero-shot-CoT, we devise another prompt in Step 2 to get the LLM to extract the final numerical answer from the reasoning text generated in Step 1.  This prompt includes the answer extraction instruction appended to the first prompt followed by the LLM generated reasoning text.  This way, LLM is expected to return the final answer in the desired form.

Based on the example in Figure 3(b), the prompt used in Step 2 will include ``\textit{Q: Grace weighs 125 pounds $\cdots$ Variables: Grace: 125 pounds $\cdots$ Answer: Combined weight of Grace and Alex = 125 + 498 = 623 pounds. \underline{Therefore, the answer (arabic numerals) is}}''. For this example, the final answer returned by LLM is ``\textit{623}''.


\section{Experimental Setup}

\subsection{Benchmarks}
The proposed method is evaluated on the ten benchmark datasets from three categories of reasoning problems: 
\textbf{Arithmetic Reasoning:} (1) the {GSM8K}~\citep{gsm8k} dataset of high quality linguistically diverse grade school math word problems created by human problem writers, (2) the {SVAMP}~\citep{svamp} benchmark of one-unknown arithmetic word problems for up-to-4 grade level students by making simple changes to a set of problems from another existing dataset, (3) the {MultiArith}~\citep{mutli_arith} dataset of math word problems requiring multiple reasoning steps and operations, (4) the {AddSub}~\cite{addsub} dataset of addition and subtraction arithmetic word problems, (5) the {AQUA}~\citep{aqua} dataset of algebraic word problems with natural language rationales, and (6) the {SingleEq}~\citep{singleeq} dataset of single-equation grade-school algebra word problems with multiple math operations over non-negative rational numbers and one variable; 
\textbf{Commonsense Reasoning}: (7) the {CSQA}~\citep{commonsenseqa} benchmark dataset of multiple-choice questions that require different types of commonsense knowledge to obtain the correct answers; and (8) the {StrategyQA}~\citep{strategyqa} benchmark dataset with questions requiring multi-step reasoning but the reasoning steps are not given. Hence, they are to be inferred; 
\textbf{Symbolic Reasoning}: (9) the {Last Letter Concatenation}~\cite{cot_wei} dataset of questions requiring the last letters of words in a name to be concatenated (e.g., ``\textit{James Brown}'' → ``\textit{sn}''), and (10) the {Coin Flip}~\cite{cot_wei} dataset of questions on whether a coin is still heads up after it is flipped or not flipped based on steps given in the questions. 
Table~\ref{tab:dataset_description} shows dataset statistics.

\begin{table}[tb]
\centering\small
\caption{Details of datasets being evaluated. Math: arithmetic reasoning. CS: commonsense reasoning. Sym.: symbolic reasoning.} 
\vspace{-5pt}
\setlength{\tabcolsep}{3pt}
\label{tab:dataset_description}
\begin{tabular}
{lcccr}
\toprule
Dataset & Domain & \# Samples &Ave. words &Answer  \\\midrule
{MultiArith} & Math &600 &31.8  &Number \\
{AddSub}  & Math &395 &31.5&Number  \\
{GSM8K}  & Math &1319 &46.9 &Number  \\
{AQUA} & Math &254 &51.9  &Option \\
{SingleEq}  & Math&508 &27.4 &Number \\
{SVAMP}   & Math &1000 &31.8 &Number \\
{CSQA}  & CS &1221 &27.8  &Option \\
{StrategyQA}  & CS &2290 &9.6  &Yes / No   \\
Last Letters  & Sym. &500 &15.0 & String  \\
Coin Flip   & Sym. &500 &37.0 & Yes / No \\
\bottomrule
\end{tabular}
\vspace{-10pt}
\end{table}

\subsection{Zero-shot and Few-shot Baselines} We compare our proposed zero-shot PS and PS+ prompting methods with three types of prompting baselines: 
(1) \textbf{Zero-shot baselines.} We include zero-shot-CoT~\cite{kojima2022large} and zero-shot-PoT~\citep{chen2022program}. The former appends ``Let's think step by step'' to the prompt without any demonstration examples. The latter uses LLM (mainly OpenAI Codex\footnote{https://openai.com/blog/openai-codex/}) to generate a Python program and then derive an answer by executing the generated program on a Python interpreter;
(2) \textbf{Few-shot with manual demonstrations.} Manual-CoT~\cite{cot_wei} creates eight hand-crafted examples as demonstrations.
(3) \textbf{Few-shot with automatic demonstrations.} Auto-CoT~\citep{zhang2022automatic} automatically selected examples by clustering with diversity and generates reasoning chains using zero-shot-CoT to construct demonstrations.

\subsection{Implementations} Following Auto-CoT~\cite{zhang2022automatic}, we use the public GPT-3~\cite{brown2020language} (175B) as the backbone language model, which is one of the most widely-used LLMs with public APIs\footnote{https://beta.openai.com/docs/models/gpt-3}. 
Since \texttt{text-davinci-003} is an upgraded version of \texttt{text-davinci-002}, which can produce higher-quality writing, accommodate more complex instructions, and perform better at longer-form content generation, We report the results using \texttt{text-davinci-003} engine for GPT-3 in the main paper.
We set the temperature to 0 (argmax sampling) throughout our experiments for the greedy decoding strategy.
We also include two few-shot baselines, Manual-CoT and Auto-CoT, we use 8 demonstration examples for MultiArith, GSM8K, AddSub, SingleEq, and SVAMP, 4 examples for AQuA and Last Letters, 7 examples for CSQA, and 6 examples for StrategyQA as suggested in the original papers, \citet{cot_wei} and \citet{zhang2022automatic}. 
Evaluation metrics wise, we follow Manual-CoT~\cite{cot_wei} and report the accuracy of all methods across datasets.


\section{Experimental Results}

\subsection{Main Results}

\begin{table*}[t]\centering


\setlength{\tabcolsep}{6pt}
\caption{Accuracy comparison on six math reasoning datasets. The best and second best results are boldfaced and underlined respectively. 
}
 \vspace{-7pt}
\small
\begin{tabular}{llccccccc}\toprule
Setting &Method (text-davinci-003)  &MultiArith   &GSM8K &AddSub &AQuA &SingleEq &SVAMP & Average \\\midrule



\multirow{ 4}{*}{Zero-Shot}& CoT & {83.8}  & {56.4}& {85.3}& {38.9} & {88.1} & {69.9} & 70.4\\

& PoT   & \bf {92.2}  & {57.0}& 85.1&\underline {43.9} & \underline{ 91.7} & {70.8} & \underline{73.5}\\



& PS (ours) &  {87.2}  & \underline{58.2}& \underline{88.1}& {42.5} & {89.2} & \underline{72.0} & 72.9\\

& PS+ (ours)  & \underline{91.8}  & \bf{59.3}& \bf 92.2& \bf{46.0} & \bf {94.7} &\bf  {75.7} & \bf 76.7\\

\midrule



\multirow{ 2}{*}{Few-Shot} & Manual-CoT & {93.6}  & {58.4}& 91.6& {48.4} & {93.5} & {80.3} & 77.6\\







& Auto-CoT & {95.5}  & {57.1}& 90.8& {41.7} & {92.1} & {78.1} & 75.9 \\

\bottomrule
\end{tabular}
\label{tab:main_results}
 \vspace{-5pt}
\end{table*}

\begin{table}[]
    \centering\small
    \setlength{\tabcolsep}{4pt}
    \caption{Accuracy on commonsense reasoning datasets.}
     \vspace{-5pt}
    \setlength{\tabcolsep}{3pt}
            \begin{tabular}{lccc}
    \toprule
     {Method}& {CSQA} & {StrategyQA}&  \\
     
    \midrule
    Few-Shot-CoT (Manual) & 78.3 & 71.2 \\
    \midrule
    Zero-shot-CoT & 65.2 & 63.8 \\
    Zero-shot-PS+ (ours) &\textbf{71.9} &  \textbf{65.4} \\
    \bottomrule
    \end{tabular}
    
    \label{tab:cs}
     \vspace{-5pt}
\end{table}
\begin{table}[]
    \centering\small
    \setlength{\tabcolsep}{4pt}
    \caption{Accuracy on symbolic reasoning datasets.}
     \vspace{-5pt}
    \setlength{\tabcolsep}{3pt}
            \begin{tabular}{lccc}
    \toprule
     {Method}& {Last Letter} & {Coin Flip}&  \\

    \midrule
    Few-Shot-CoT (Manual) & 70.6 & 100.0 \\
    \midrule
    Zero-shot-CoT & 64.8 & 96.8 \\
    Zero-shot-PS+ (ours) &\textbf{75.2} &  \bf 99.6 \\
    \bottomrule
    \end{tabular}
    
    \label{tab:sr}
     \vspace{-5pt}
\end{table}

\paragraph{Arithmetic Reasoning.} Table~\ref{tab:main_results} reports the accuracy comparison of our method and existing zero-shot and few-shot methods on the arithmetic reasoning datasets. 
In the zero-shot setting, our PS+ prompting (i.e., PS prompting with more detailed instructions) consistently outperforms Zero-shot-CoT across all arithmetic reasoning datasets by a large margin.
Specifically, PS+ prompting improves the accuracy over Zero-shot CoT by at least 5\% for all datasets except GSM8K which sees a 2.9\% improvement. 
The exception could be due to GSM8K being a more challenging dataset from the linguistics complexity aspect. 
PS prompting also outperforms Zero-shot-CoT across all datasets, and enjoys 2.5\% higher average accuracy than that of Zero-shot CoT. 

Compared with another competitive Zero-shot baseline, PoT, the performance of PS(+) and PS promptings are still impressive. PS+ prompting outperforms PoT on five out of six arithmetic datasets. PS prompting also outperforms PoT on three arithmetic datasets.
The results suggest that adding more detailed instructions to the prompt can effectively elicit higher-quality reasoning steps from LLMs.

Compared with the few-shot methods, Manual CoT and Auto-CoT, PS+ prompting yields an average accuracy (76.7\%) slightly lower than Manual-CoT (77.6\%) but higher than Auto-CoT (75.9\%).
While this is an unfair comparison, this result indicates that zero-shot prompting can outperform few-shot CoT prompting, which hopefully will spark further development of new ways with a less manual effort to effectively elicit reasoning in LLMs.

\paragraph{Commmonsense Reasoning.}

Table \ref{tab:cs} shows the results on commonsense reasoning datasets: CommonsenseQA and StrategyQA. We only include our better zero-shot PS+ prompting strategy in this comparison.  Zero-shot PoT is excluded as it does not work on this problem. 
While PS+ prompting underperforms Few-Shot-CoT(Manual) on this problem, it consistently outperforms Zero-shot-CoT on CommonsenseQA (71.9\% vs. 65.2\%) and StrategyQA (65.4\% vs. 63.8\%) datasets.

\paragraph{Symbolic Reasoning.}
Table~\ref{tab:sr} shows the accuracy of PS+ prompting against Zero-shot-CoT and Few-shot-CoT on symbolic reasoning datasets: Last Letters and Coin Flip. Zero-shot PoT is again excluded as it is not designed for the problem.
On Last Letters, our Zero-shot PS+ prompting (75.2\%) outperforms Manual-CoT (70.6\%) and Zero-shot-CoT (65.2\%). On Coin Flip, Zero-shot PS+ prompting (99.6\%) is slightly worse than Manual-CoT (100.0\%) but outperforms Zero-shot-CoT by a good margin (96.8\%).
More examples from the experiment results can be found in Appendix~\ref{sec:appendix_examples}.

\subsection{Analysis}

\begin{table*}[t]\centering
\caption{Performance comparison of trigger sentences measured on GSM8K and SVAMP datasets with \texttt{text-davinci-003} except for No. 2 (\texttt{code-davinci-002}). (*1) means the trigger sentence used in Zero-shot-CoT~\cite{kojima2022large}. (*2) means the trigger sentence used in Zero-shot-PoT~\cite{chen2022program}.}
 \vspace{-5pt}
\small
\begin{tabular}{lllcc}\toprule
No. &Trigger Sentence&GSM8K&SVAMP \\
\midrule
1 &\makecell[l]{Let's think step by step.} (*1) &56.4  & 69.9 \\
\midrule

2 & \makecell[l]{import math \\
import numpy as np \\
\# Question: {example['question']} \\
\# Answer this question by implementing a solver()  function. \\
def solver():\\
\qquad \# Let's write a Python program step by step, and then return the answer \\
\qquad  \# Firstly, we need define the following variable:
} (*2)
 &57.0 & 70.8 \\
\midrule

3&\makecell[l]{{\sethlcolor{light_red}\hl{Extract variables}} and assign {\sethlcolor{light_red}\hl{their  corresponding  numerals }} to  these variables\\ first and  then solve  the problem step by step.} &50.5  & 69.5 \\
\midrule
4&\makecell[l]{Firstly, {\sethlcolor{light_red}\hl{extract variables}} and {\sethlcolor{light_red}\hl{their corresponding numerals}}.  Then, {\sethlcolor{light_red}\hl{calculate}} \\{\sethlcolor{light_red}\hl{intermediate variables}}. Finally, solve the    problem step by step.} &54.8  & 70.8 \\

\midrule
5&\makecell[l]{Let's first understand the problem and {\sethlcolor{light_blue}\hl{devise a plan}} to solve  the problem. \\
Then, let's {\sethlcolor{light_blue}\hl{carry out the plan}}  and solve the problem step by step.} &58.2 & 72.0 \\

\midrule
6&\makecell[l]{Let's first understand the problem, {\sethlcolor{light_red}\hl{extract relevant variables}}  and {\sethlcolor{light_red}\hl{their}} \\ {\sethlcolor{light_red}\hl{corresponding numerals}}, and {\sethlcolor{light_blue}\hl{make a plan}}. Then,  let's {\sethlcolor{light_blue}\hl{carry out the plan}},\\ {\sethlcolor{light_red}\hl{calculate intermediate variables (pay attention to correct numerical}} \\ {\sethlcolor{light_red}\hl{ calculation and commonsense)}},  solve the problem step by step, and show\\ the answer. } &\bf 59.3 & \bf 75.7 \\






\bottomrule
\end{tabular}
\label{tab:template_study}
 \vspace{-5pt}
\end{table*}

\begin{figure}[t]

  \vspace{-10pt}
  \centering
  
  \includegraphics[width=1.0\linewidth]{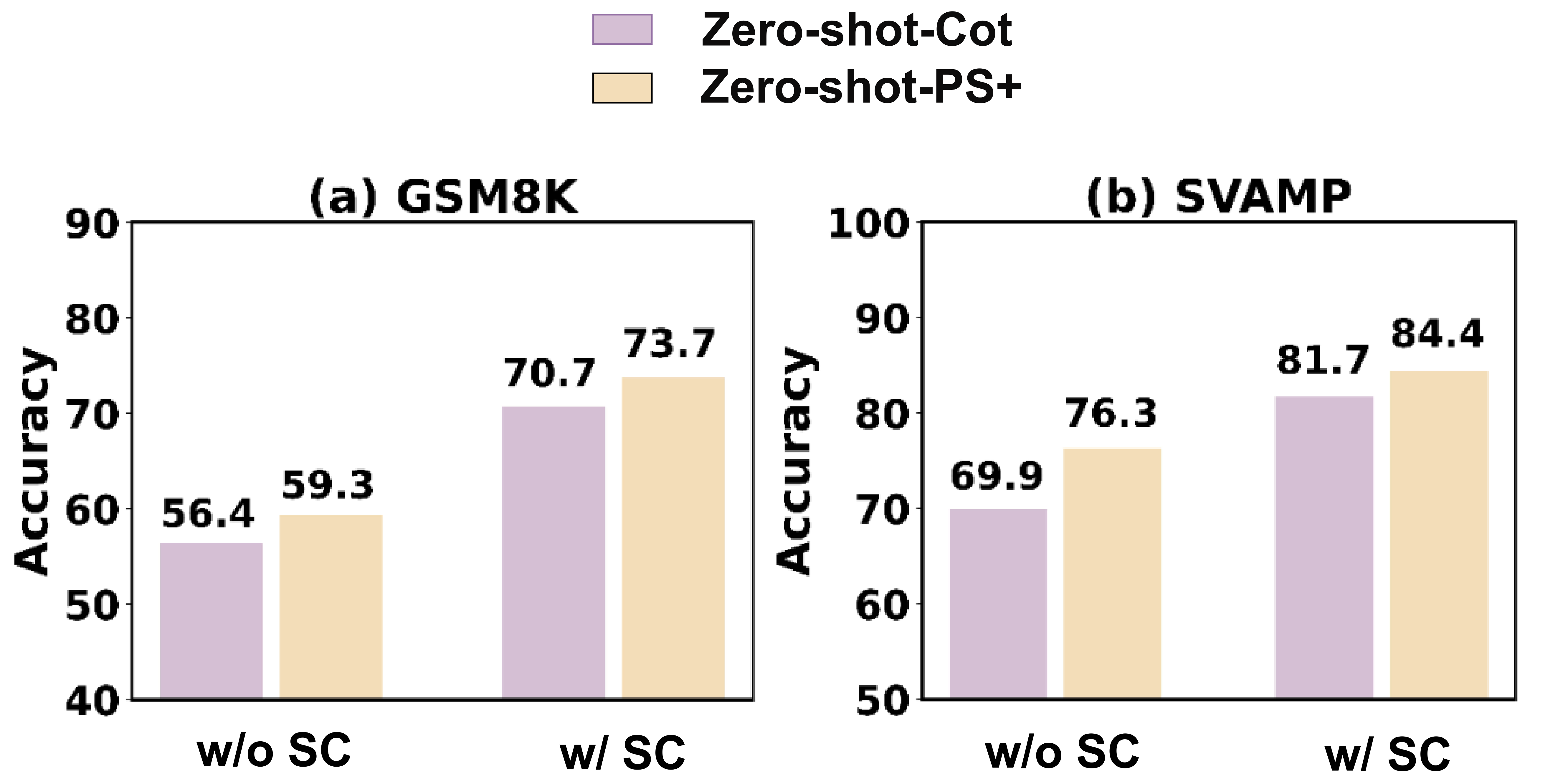}
  \vspace{-10pt}
  \caption{
    Results of methods with and without self-consistency (SC) on GSM8K and SVAMP. 
  }
  
  \label{fig:sc}
\vspace{-10pt}
\end{figure}

\paragraph{Results of Prompting with Self-Consistency.}
Self-consistency~\citep{wang2022self_consistency} (SC) is proposed to reduce randomness in LLM's output by generating $N$ reasoning results and determining the final answer by majority voting.  With SC, the methods' results are usually expected to be consistent and better.
Hence, we evaluate Zero-shot PS+ prompting with SC on GSM8K and SVAMP datasets.
We set the temperature to 0.7 and $N$ to 10 for experiments with SC. 
Figure~\ref{fig:sc} shows that PS+ prompting with SC (73.7\% and 84.4\%) substantially outperforms that without SC (58.7\% and 75.7\%) on GSM8K and SVAMP, respectively.  The former also consistently outperforms Zero-shot-CoT with SC (70.7\% and 81.7\%) on GSM8K and SVAMP, respectively, although Zero-shot CoT also enjoys improvement with the self consistency approach.

\paragraph{Effect of Prompts. }
Table~\ref{tab:template_study} demonstrates a comparison of the performance of 6 different input prompts. Prompts 1 and 2 are used in Zero-shot CoT and Zero-shot PoT respectively.  The rest are variations of prompts used in Step 1 of the Zero-shot PS+ prompting strategies with greedy decoding.
We observe that Prompt 3 with variables and numeral extraction performs worse than Prompt 1 of Zero-shot-CoT. 
The reason is that Prompt 3 doesn't include instructions for devising and completing a plan. 
However, the other prompts of Zero-shot-PS+ perform well as we add more instructions about intermediate results calculation, plan design, and implementation.
The above results conclude that LLMs are capable of generating high-quality reasoning text when the prompts include more detailed instructions to guide the LLMs.
More prompts for different reasoning problems can be found in Appendix~\ref{sec:appendix_results}.

\paragraph{Error Analysis.} To qualitatively evaluate the impact of the Zero-shot-PS+ prompting on calculation errors and reasoning steps missing errors, we examine the distribution of errors on the GSM8K dataset. We first randomly sample 100 problems from GSM8K, generate the reasoning text, and extract answers using Zero-Shot-CoT, Zero-shot-PS, and Zero-shot-PS+ prompting strategies. Zero-Shot-CoT generated incorrect final answers for 46 of the problems, 43 for Zero-shot-PS, and 39 for Zero-shot-PS+. Subsequently, we analyze and determine the error types of all these problems as shown in Table~\ref{tab:error}.

\begin{table}[]
    \caption{Distribution of error types of 100 examples from GSM8K where Zero-shot-CoT, zero-shot PS (Zero-shot-PS) prompting, and zero-shot PS+ prompting get incorrect final answers.}
     \vspace{-5pt}
    \centering\small
    \setlength{\tabcolsep}{4pt}
    \setlength{\tabcolsep}{3pt}
    \begin{tabular}{lccc}
    \toprule
     \textbf{Method}& \textbf{Calculation} & \textbf{Missing}& \textbf{Semantic}  \\
     
    \midrule
    Zero-shot-CoT & 7\% & 12\% & 27\% \\
    \midrule
    Zero-shot-PS & 7\% & 10\% & 26\%  \\
    Zero-shot-PS+  & 5\% & 7\% & 27\% \\
    \bottomrule
    \end{tabular}

    \label{tab:error}
    \vspace{-10pt}
\end{table}

The analysis results show that PS+ prompting achieves the least calculation (5\%) and missing-step  (7\%) errors, and semantic understanding errors comparable to Zero-shot-CoT.  Zero-shot-PS has slightly more errors but is still better than Zero-shot-CoT.  Their plan-and-solve prompts thus effectively guide the LLMs to generate clear and complete reasoning steps. Moreover, the additional detailed instructions in PS+ prompting (i.e., ``\textit{extract relevant variables and their corresponding numerals}'' and ``\textit{calculate intermediate variables}'') enable the LLMs to generate high-quality reasoning steps leading to fewer calculation errors. 

\begin{figure}[t]
  \centering
  \includegraphics[width=0.78\linewidth]{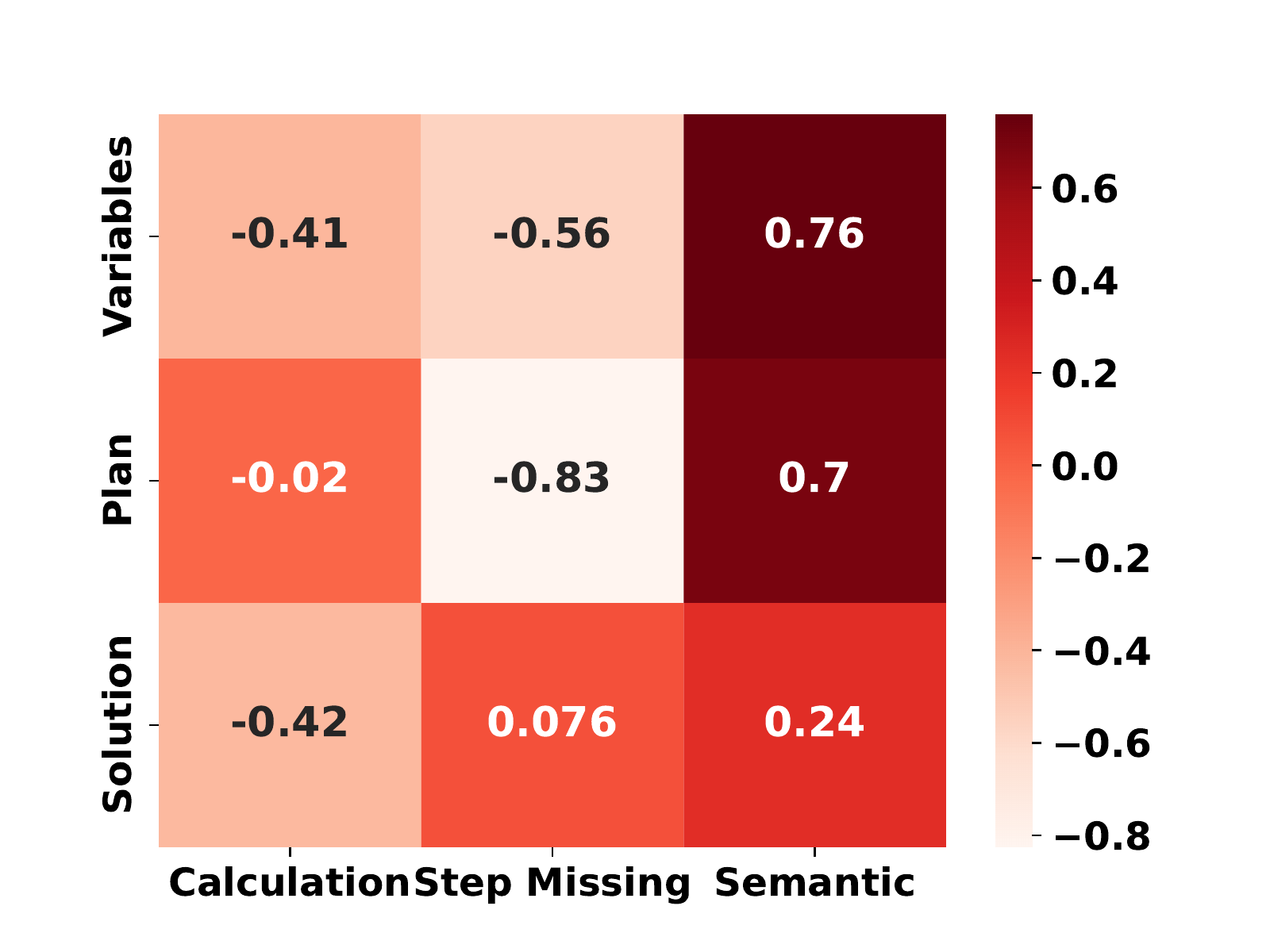}
  \vspace{-10pt}
  \caption{
    Correlation analysis of generated reasoning and error types of randomly sampled 100 data examples from GSM8K for Zero-shot-PS+.
  }
  \label{fig:correlation}
\vspace{-10pt}
\end{figure}

\paragraph{Correlation Analysis of Generated Reasoning and Error Types.}
To obtain deeper insight into the impact of PS+ prompting on error types, we examine the correlation between the sub-parts of the generated reasoning and error types. Specifically, we analyze the existence of variable definition, reasoning plan, and solution in the generated reasoning text and correlate them with the three error types.  The set of problems used for this analysis study is the same as that used in the earlier error type analysis. 
Figure \ref{fig:correlation} shows the correlation matrix among the existence of variable definitions, plans, solutions and three different types of errors. It is observed that both variable definition and plan existences have a negative correlation with calculation errors and missing-reasoning-step errors. The Zero-shot-PS+ prompt can further improve the performance of LLMs on mathematical reasoning problems by reducing calculation errors and missing-reasoning-step errors.

\paragraph{Exploring the Presence of Plans in PS Predictions.}
To ascertain the presence of a plan in each prediction made by PS, we conducted a random sampling of 100 data examples and examined their corresponding predictions. Our analysis reveals that 90 of the 100 predictions indeed incorporated a plan. 
This observation indicates the emergence of strong planning abilities in recent LLMs such as GPT-3.5 and GPT-4.

\section{Related Work}
\label{sec:related_work}

\subsection{Reasoning in NLP}
It is well known that complex reasoning problems are challenging for NLP models, and such problems include mathematical reasoning~\citep{gsm8k, svamp, aqua, mawps} (requiring the ability to understand mathematical concepts, calculation, and multi-step reasoning), commonsense reasoning~\citep{commonsenseqa, strategyqa} (requiring the ability to make judgments based on commonsense knowledge), and logical reasoning~\cite{cot_wei} (requiring the ability to manipulate symbols by applying formal logical rules).
Before the advent of Large Language models (LLMs), ~\citet{commonsenseqa} trained the NLP model using explanations generated by the fine-tuned GPT model and found that the trained model yields better performance on commonsense QA problems.
~\citet{hendrycks2021measuring} attempted to fine-tune pretrained language models with labeled rationale, but found out that these fine-tuned models could not easily generate high-quality reasoning steps.
Recent work by \citet{wei2022emergent} showed that LLMs demonstrates strong reasoning ability when scaled up to tens of billions of parameters, such as GPT-3~\citep{brown2020language} and PaLM~\citep{palm}.
These LLMs
with a few demonstration exemplars can yield impressive performance across different NLP tasks. 
However, these models still perform poorly in problems that require multi-step reasoning.
This may be due to the fact that the few exemplars provided are insufficient to unlock the LLMs' capabilities.

\subsection{Prompting Methods}
To exploit the reasoning ability in LLMs, ~\citet{cot_wei} propose Chain-of-Thought prompting, appending multiple reasoning steps before the answer to the input question. 
With this simple few-shot prompting strategy, LLMs are able to perform much better in complex reasoning problems.
Subsequently, many works
\citep{wang2022towards, suzgun2022challenging, shaikh2022second_ana, saparov2022language_ana} propose to further improve CoT prompting in different aspects, including prompt format~\cite{chen2022program}, prompt selection~\citep{lu2022dynamic}, prompt ensemble~\cite{wang2022self_consistency, li2022advance, self_verification, fu2022complexity}, problem decomposition~\cite{zhou2022least_to_most, khot2022decomposed, dua2022successive, press2022measuring}, 
and planning~\cite{Yao2022ReActSR, huang2022language, wang2023describe, liu2023llm+, Sun2023PEARLPL, Yao2023TreeOT}.
\citet{chen2022program} introduced PoT prompting to use LLMs with code pre-training to write a program as a rationale for disentangling computation from reasoning.
To do away with manual effort, ~\citet{kojima2022large} proposed Zero-shot-CoT to elicit reasoning step generation without exemplars.  
To leverage the benefit of demonstration examples and minimize manual effort, ~\citet{zhang2022automatic} designed Auto-CoT. It first automatically obtains $k$ examples by clustering the given dataset. It then follows Zero-shot-CoT to generate rationales for the selected examples. Finally, demonstration examples are constructed by adding the generated rationales to selected examples as CoT prompts. 
Our work is different from the above works by focusing on eliciting multi-step reasoning by  LLMs in a zero-shot approach. We ask LLMs to write a plan to decompose a complex reasoning task into multiple reasoning steps. Furthermore, we introduce detailed instructions to the prompt to avoid obvious errors in the reasoning steps.
We refer readers to the survey~\cite{ huang2022towards_survey} for more related works.

\section{Conclusion}


In this paper, we find that Zero-shot-CoT still suffers from three pitfalls: calculation errors,  missing-reasoning-step errors, and semantic understanding errors.
To address these issues, we introduce plan-and-solve prompting strategies (PS and PS+ prompting). They are new zero-shot prompting methods that guide LLMs to devise a plan that divides the entire task into smaller subtasks and then carries out the subtasks according to the plan.
Evaluation on ten datasets across three types of reasoning problems shows PS+ prompting outperforms the previous zero-shot baselines and performs on par with few-shot CoT prompting on multiple arithmetic reasoning datasets. 
Overall, our results suggest that (a) Zero-shot PS+ prompting can generate a high-quality reasoning process than Zero-shot-CoT prompting since the PS prompts can provide more detailed instructions guiding the LLMs to perform correct reasoning; (b) Zero-shot PS+ prompting has the potential to outperform manual Few-shot CoT prompting, which hopefully will spark further development of new CoT prompting approaches to elicit reasoning in LLMs. 
Moreover, PS(+) prompting is a general idea that can be used for non-reasoning tasks, and refining the plan is also an interesting idea. We leave them for future work.

\section{Limitations}

There are two limitations to this work. First, it takes effort to design the prompt to guide the LLMs to generate correct reasoning steps. The GPT-3 models are sensitive to the expressions in prompts. Thus we need to carefully design the prompts. Second, the proposed plan-and-solve prompting can help address the calculation errors and missing-reasoning-step errors, but the semantic misunderstanding errors still remain. We will explore how to address semantic misunderstanding errors by prompting instead of upgrading LLMs in the future.

\section{Ethics}
We experiment on six math reasoning datasets, including AQuA~\cite{aqua}, GSM8K~\cite{gsm8k}, MultiArith, AddSub, SingleEq, and SVAMP~\cite{svamp}, two commonsense reasoning tasks (CommonsenseQA~\cite{commonsenseqa} and StrategyQA~\cite{strategyqa}), and two symbolic tasks (Last Letter and Coin Flip~\cite{cot_wei}), where GSM8K and SVAMP use the MIT License code, AQUA and StrategyQA use the Apache-2.0 code, the remaining datasets are unspecified. 

The proposed prompts do not collect and use personal information about other individuals. The prompts we used are listed in Appendix. The prompts in this work do not contain any words that discriminate against any individual or group. In this work, prompts would not negatively impact other people's safety.

\bibliography{custom}
\bibliographystyle{acl_natbib}

\appendix

\section{Appendix}
\label{sec:appendix}

This section includes two parts: (1) Results of all prompts we have tried; (2) Example texts generated by Zero-shot-PS+. Unless otherwise mentioned, we use GPT3 (text-davinci-003) model.


\subsection{Results of All Trigger Sentences}
\label{sec:appendix_results}

Tables 7 to 16 list the results of all prompts we have tried for each dataset.

\begin{table*}[t]\centering
\caption{Performance comparison of prompts used in Step 1 of Zero-shot-PS+ prompting with \texttt{text-davinci-003} on AQuA.}
\small
\begin{tabular}{lllc}\toprule
No. &Trigger Setence&Accuracy \\
\midrule

1 &\makecell[l]{Let's first understand the problem and devise a plan to solve the problem. 
Then, let's carry out  the plan \\to solve the problem step by step.} & 42.5  \\
\midrule
2 &\makecell[l]{Let's first understand the problem, extract all relevant  variables and their corresponding numerals \\carefully, and devise a plan. Then, let's carry out the plan, calculate intermediate variables (pay attention \\to correct numerical calculation and common sense), solve the problem step by step carefully, and show \\the answer. } & 42.9 \\
\midrule
3 &\makecell[l]{Let's first understand the problem, extract relevant correct variables and their correct corresponding \\numerals, and devise complete plans. Then, let's carry out the plan, calculate intermediate variables \\including extracted variables (pay attention to correct numerical calculation and common sense), solve \\the problem by single equations, and show the answer.} & 43.7  \\
\midrule
4 &\makecell[l]{Let's first understand the problem, extract relevant variables and  their corresponding numerals, and \\ make a complete plan.Then,  let's carry out the plan, calculate intermediate variables (pay attention to \\ correct numerical calculation and commonsense), solve the problem step by step, and show the answer.} & 46.0  \\

\bottomrule
\end{tabular}
\label{tab:p_aqua}
\end{table*}
\begin{table*}[t]\centering
\caption{Performance comparison of prompts used in Step 1 of Zero-shot-PS+ prompting with \texttt{text-davinci-003} on GSM8K.}
\small
\begin{tabular}{lllc}\toprule
No. &Trigger Setence&Accuracy \\
\midrule

1 &\makecell[l]{Let's first understand the problem and devise a plan to solve the problem. 
Then, let's carry out the plan \\ to solve the problem step by step.} & 58.2  \\
\midrule
2 &\makecell[l]{Let's first understand the problem, extract relevant variables and their corresponding numerals, and \\devise a plan.
Then, let's carry out the plan, calculate intermediate variables (pay attention to correct \\numeral calculation and commonsense), solve the problem step by step, and show the answer. } &  58.7\\
\midrule
3 &\makecell[l]{Let's first understand the problem, extract relevant variables and  their corresponding numerals, and\\ make a complete plan.Then,  let's carry out the plan, calculate intermediate variables (pay attention to \\correct numerical calculation and commonsense), solve the problem step by step, and show the answer.} & 59.3 \\

\bottomrule
\end{tabular}
\label{tab:p_aqua}
\end{table*}
\begin{table*}[t]\centering
\caption{Performance comparison of prompts used in Step 1 of Zero-shot-PS+ prompting with \texttt{text-davinci-003} on MultiArith.}
\small
\begin{tabular}{lllc}\toprule
No. &Trigger Setence&Accuracy \\
\midrule

1 &\makecell[l]{Let's first understand the problem and devise a plan to solve the problem. 
Then, let's carry out the plan \\to solve the problem step by step.} & 87.2  \\
\midrule
2 &\makecell[l]{Let's first understand the problem, extract relevant variables and their corresponding numerals, and \\devise a plan.
Then, let's carry out the plan, calculate intermediate variables (pay attention to correct \\numeral calculation and commonsense), solve the problem step by step, and show the answer. } & 88.3 \\
\midrule
3 &\makecell[l]{Let's first understand the problem, extract relevant variables and their corresponding numerals, and \\devise a complete plan. Then, let's carry out the plan, calculate intermediate variables (pay attention \\to the correctness of the calculation and common sense), solve the problem step by step, and show \\the answer.} & 90.5 \\
\midrule
4 &\makecell[l]{Let's first understand the problem, extract relevant variables and their corresponding numerals, and \\devise a complete plan. Then, let's carry out the plan, calculate intermediate variables (pay attention to \\correct numerical calculation and commonsense), solve the problem step by step, and show the answer.} &  91.8 \\

\bottomrule
\end{tabular}
\label{tab:p_aqua}
\end{table*}
\begin{table*}[t]\centering
\caption{Performance comparison of prompts used in Step 1 of Zero-shot-PS+ prompting with \texttt{text-davinci-003} on SVAMP.}
\small
\begin{tabular}{lllc}\toprule
No. &Trigger Setence&Accuracy \\
\midrule

1 &\makecell[l]{Let's first understand the problem and devise a plan to solve the problem. 
Then, let's carry out the plan \\to solve the problem step by step.} & 72.0  \\
\midrule
2 &\makecell[l]{Let's first understand the problem, extract relevant variables and their corresponding numerals, and \\devise a plan.
Then, let's carry out the plan, calculate intermediate variables (pay attention to correct\\ numeral calculation and commonsense), solve the problem step by step, and show the answer. } & 75.4  \\
\midrule
3 &\makecell[l]{Let's first understand the problem, extract relevant variables and  their corresponding numerals, and\\ make a complete plan.Then,  let's carry out the plan, calculate intermediate variables (pay attention to \\correct numerical calculation and commonsense), solve the problem step by step, and show the answer.} & 75.7 \\

\bottomrule
\end{tabular}
\label{tab:p_aqua}
\end{table*}
\begin{table*}[t]\centering
\caption{Performance comparison of prompts used in Step 1 of Zero-shot-PS+ prompting with \texttt{text-davinci-003} on AddSub.}
\small
\begin{tabular}{lllc}\toprule
No. &Trigger Setence&Accuracy \\
\midrule

1 &\makecell[l]{Let's first understand the problem and devise a plan to solve the problem. 
Then, let's carry out the plan \\ to solve the problem step by step.} & 87.3  \\
\midrule
2 &\makecell[l]{Let's first understand the problem, extract relevant variables and their corresponding numerals, and \\devise a complete plan.Then, let's carry out the plan, calculate intermediate variables (pay attention to \\correct numerical calculation and commonsense), solve the problem step by step, and show the answer.} & 87.8 \\
\midrule
3 &\makecell[l]{Let's first understand the problem, extract relevant variables and their corresponding numerals, and\\ devise a plan.
Then, let's carry out the plan, calculate intermediate variables (pay attention to correct \\ numeral calculation and commonsense), solve the problem step by step, and show the answer.} & 92.2 \\

\bottomrule
\end{tabular}
\label{tab:p_aqua}
\end{table*}
\begin{table*}[t]\centering
\caption{Performance comparison of prompts used in Step 1 of Zero-shot-PS+ prompting with \texttt{text-davinci-003} on SingleEq.}
\small
\begin{tabular}{lllc}\toprule
No. &Trigger Setence&Accuracy \\
\midrule

1 &\makecell[l]{Let's first understand the problem and devise a plan to solve the problem. 
Then, let's carry out the plan\\ to solve the problem step by step.} & 92.3  \\
\midrule
2 &\makecell[l]{Let's first understand the problem, extract relevant variables and their corresponding numerals, and \\devise a plan.
Then, let's carry out the plan, calculate intermediate variables (pay attention to correct\\ numeral calculation and commonsense), solve the problem step by step, and show the answer. } & 94.7 \\

\bottomrule
\end{tabular}
\label{tab:p_aqua}
\end{table*}
\begin{table*}[t]\centering
\caption{Performance comparison of prompts used in Step 1 of Zero-shot-PS+ prompting with \texttt{text-davinci-003} on CSQA.}
\small
\begin{tabular}{lllc}\toprule
No. &Trigger Setence&Accuracy \\
\midrule

1 &\makecell[l]{Let's devise a plan and solve the problem step by step.} & 67.4\\
\midrule
2 &\makecell[l]{Let's first understand the problem, extract relevant variables and their corresponding numerals, and \\devise a complete plan.Then, let's carry out the plan, calculate intermediate variables (pay attention to \\correct numerical calculation and commonsense), solve the problem step by step, and show the answer.} & 71.9 \\

\bottomrule
\end{tabular}
\label{tab:p_aqua}
\end{table*}
\begin{table*}[t]\centering
\caption{Performance comparison of prompts used in Step 1 of Zero-shot-PS+ prompting with \texttt{text-davinci-003} on StrategyQA.}
\small
\begin{tabular}{lllc}\toprule
No. &Trigger Setence&Accuracy \\
\midrule

1 &\makecell[l]{Let's devise a plan and solve the problem step by step.} & 61.5  \\
\midrule
2 &\makecell[l]{Let's devise a complete plan. Then, let's carry out the plan, solve the problem step by step, and \\show the answer.} & 63.0 \\
\midrule
3 &\makecell[l]{Let's first prepare relevant information and make a plan. Then, let's answer the question step by step \\(pay attention to commonsense and logical coherence).} &  65.4\\

\bottomrule
\end{tabular}
\label{tab:p_aqua}
\end{table*}
\begin{table*}[t]\centering
\caption{Performance comparison of prompts used in Step 1 of Zero-shot-PS+ prompting with \texttt{text-davinci-003} on Last Letters.}
\small
\begin{tabular}{lllc}\toprule
No. &Trigger Setence&Accuracy \\
\midrule

1 &\makecell[l]{Let's devise a plan and solve the problem step by step.} & 75.2  \\

\bottomrule
\end{tabular}
\label{tab:p_aqua}
\end{table*}
\begin{table*}[t]\centering
\caption{Performance comparison of prompts used in Step 1 of Zero-shot-PS+ prompting with \texttt{text-davinci-003} on Coin Flip.}
\small
\begin{tabular}{lllc}\toprule
No. &Trigger Setence&Accuracy \\
\midrule

1 &\makecell[l]{Let's devise a complete plan. Then, let's carry out the plan, solve the problem step by step,\\ and show the answer.} & 70.6 \\
\midrule
2 &\makecell[l]{Let's first devise a plan, then solve the problem step by step.} & 72.6  \\
\midrule
3 &\makecell[l]{Let's first devise a plan, then solve the problem step by step.(Distinguish between tail up and \\ head up)} & 84.0 \\
\midrule
4 &\makecell[l]{Let's first understand the problem and devise a complete plan.Then, let's carry out the plan, \\reason problem step by step. Every step answer the subquestion "does the person flip and \\what is the coin current state?", and according to the last coin state, give the final answer \\(pay attention that the coin turns state at every flip). \\
Plan: \\
Step 1:} & 94.2 \\
\midrule
5 &\makecell[l]{Let's first understand the problem and devise a complete plan. Then, let's carry out the plan, \\reason problem step by step. Every step answer the subquestion "does the person flip and what \\is the coin current state?", and according to the coin last state give the final answer(pay attention \\to every flip, the coin turns state). \\
Plan: \\
Step 1:} & 99.0 \\
\midrule
6 &\makecell[l]{Let's first understand the problem and devise a complete plan. Then, let's carry out the plan and \\reason problem step by step. Every step answer the subquestion, "does the person flip and what is\\ the coin's current state?". According to the coin's last state, give the final answer (pay attention to \\every flip and the coin’s turning state).} & 99.6 \\

\bottomrule
\end{tabular}
\label{tab:p_aqua}
\end{table*}

\subsection{Example Outputs by Zero-shot-PS+}
\label{sec:appendix_examples}

Tables 17 to 25 list example outputs generated by Zero-shot-PS+ for each dataset.

\begingroup
\begin{table*}[htb]
    \centering
    \caption{
    Example outputs by Zero-shot-PS+ for AQuA. 
    }
    \vspace{2.8mm}
    \begin{tabular}{p{0.96\linewidth}}
        \toprule
        
\textbf{Q:} The average wages of a worker during a fortnight comprising of 15 consecutive working days was $\$90$ per day. During the first $7$ days, his average wage was $\$87$ per day and the average wage during the last $7$ days was $\$92$ per day. What was his wage on the $8$th day? Answer Choices: (A) $\$83$ (B) $\$92$ (C) $\$90$ (D) $\$97$ (E) $\$104$\\
\vspace{-1mm}
\textbf{A:} Let's first understand the problem, extract relevant variables and  their corresponding numerals, and make a complete plan.Then,  let's carry out the plan, calculate intermediate variables (pay attention to correct numerical calculation and commonsense), solve the problem step by step, and show the answer.\\
\vspace{-1mm}
\textbf{Output:}\\
Variables:\\
Average wage during the first 7 days = 87\\
Average wage during the last 7 days = 92\\
Average wage during the fortnight = 90\\
Plan:\\
Calculate the total wages earned during\\ the first 7 days.\\
Calculate the total wages earned during\\ the last 7 days.\\
Calculate the total wages earned during\\ the fortnight.\\
Calculate the wage on the 8th day.\\
Calculation:\\
Total wages earned during the first 7 days = 7 x 87 = 609\\
Total wages earned during the last 7 days = 7 x 92 = 644\\
Total wages earned during the fortnight = 15 x 90 = 1350\\
Wage on the 8th day = (1350 - 609 - 644) / 1 = 97\\
Answer: (D) $\$97$.\\
\midrule
\textbf{Q:} If 27 bottles of soda cost A cents, how much will B bottles cost in dollars? Answer Choices: (A) AB/2700 (B) 27/AB (C) AB/270 (D) 2700/AB (E) 100AB/27\\
\vspace{-1mm}
\textbf{A:} Let's first understand the problem, extract relevant variables and  their corresponding numerals, and make a complete plan.Then,  let's carry out the plan, calculate intermediate variables (pay attention to correct numerical calculation and commonsense), solve the problem step by step, and show the answer.\\
\vspace{-1mm}
\textbf{Output:}\\
Given: \\
27 bottles of soda cost A cents\\
B bottles of soda\\
Plan:\\
1. Convert A cents to dollars\\
2. Calculate the cost of B bottles of soda\\
Solution:\\
1. Convert A cents to dollars: A cents = A/100 dollars\\
2. Calculate the cost of B bottles of soda: B bottles of soda = (A/100) * (B/27) = AB/2700 dollars\\
Answer: (A) AB/2700\\

\bottomrule
    \end{tabular}
\end{table*}
\endgroup
\begingroup
\begin{table*}[htb]
    \centering
    \caption{
    Example outputs by Zero-shot-PS+ for GSM8K. 
    }
    \vspace{2.8mm}
    \begin{tabular}{p{0.96\linewidth}}
        \toprule
        
\textbf{Q:} James decides to run 3 sprints 3 times a week. He runs 60 meters each sprint. How many total meters does he run a week?\\
\vspace{-1mm}
\textbf{A:} Let's first understand the problem, extract relevant variables and their corresponding numerals, and make a complete plan.Then, let's carry out the plan, calculate intermediate variables (pay attention to correct numerical calculation and commonsense), solve the problem step by step, and show the answer.\\
\vspace{-1mm}
\textbf{Output:}\\
Given:\\
James runs 3 sprints 3 times a week.\\
Each sprint is 60 meters.\\
Plan:\\
We need to calculate the total meters run by James in a week.\\
Calculation:\\
Total number of sprints run by James in a week = 3 sprints x 3 times = 9 sprints\\
Total meters run by James in a week = 9 sprints x 60 meters = 540 meters\\
Answer:\\
James runs 540 meters in a week.\\
\midrule
\textbf{Q:} In a dance class of 20 students, 20\% enrolled in contemporary dance, 25\% of the remaining enrolled in jazz dance, and the rest enrolled in hip-hop dance. What percentage of the entire students enrolled in hip-hop dance?\\
\vspace{-1mm}
\textbf{A:} Let's first understand the problem, extract relevant variables and their corresponding numerals, and make a complete plan.Then, let's carry out the plan, calculate intermediate variables (pay attention to correct numerical calculation and commonsense), solve the problem step by step, and show the answer.
\\
\vspace{-1mm}
\textbf{Output:}\\
Given:\\
Total number of students = 20\\
Percentage of students enrolled in contemporary dance = 20\%\\
Percentage of students enrolled in jazz dance = 25\%\\
Plan:\\
1. Calculate the number of students enrolled in contemporary dance.\\
2. Calculate the number of students remaining after subtracting the number of students enrolled in contemporary dance.\\
3. Calculate the number of students enrolled in jazz dance.\\
4. Calculate the number of students enrolled in hip-hop dance.\\
5. Calculate the percentage of students enrolled in hip-hop dance.\\
Calculation:\\
1. Number of students enrolled in contemporary dance = 20\% of 20 = 20\% × 20 = 4\\
2. Number of students remaining after subtracting the number of students enrolled in contemporary dance = 20 - 4 = 16\\
3. Number of students enrolled in jazz dance = 25\% of 16 = 25\% × 16 = 4\\
4. Number of students enrolled in hip-hop dance = 16 - 4 = 12\\
5. Percentage of students enrolled in hip-hop dance = 12/20 × 100\% = 60\%\\
Answer:\\
60\% of the entire students enrolled in hip-hop dance.\\
\bottomrule
    \end{tabular}
\end{table*}
\endgroup
\begingroup
\begin{table*}[htb]
    \centering
    \caption{
    Example outputs by Zero-shot-PS+ for MultiArith. 
    }
    \vspace{2.8mm}
    \begin{tabular}{p{0.96\linewidth}}
        \toprule
        
\textbf{Q:} The school cafeteria ordered 42 red apples and 7 green apples for students lunches. But, if only 9 students wanted fruit, how many extra did the cafeteria end up with? \\
\vspace{-1mm}
\textbf{A:} Let's first understand the problem, extract relevant variables and their corresponding numerals, and devise a complete plan.Then, let's carry out the plan, calculate intermediate variables (pay attention to correct numerical calculation and commonsense), solve the problem step by step, and show the answer.
\\
\vspace{-1mm}
\textbf{Output:}\\
Variables: \\
Red apples = 42 \\
Green apples = 7 \\
Students wanting fruit = 9 \\
Plan: \\
Calculate the total number of apples ordered by the cafeteria \\
Subtract the total number of apples ordered by the cafeteria from the number of students wanting fruit \\
Calculation: \\
Total number of apples ordered by the cafeteria = 42 + 7 = 49 \\
Answer: \\
Extra apples = 49 - 9 = 40\\
\midrule
\textbf{Q:} Carol and her mom were picking carrots from their garden. Carol picked 29 and her mother picked 16. If only 38 of the carrots were good, how many bad carrots did they have? \\
\vspace{-1mm}
\textbf{A:} Let's first understand the problem, extract relevant variables and their corresponding numerals, and devise a complete plan.Then, let's carry out the plan, calculate intermediate variables (pay attention to correct numerical calculation and commonsense), solve the problem step by step, and show the answer.
\\
\vspace{-1mm}
\textbf{Output:}\\
Variables: \\
Carol: 29 \\
Mom: 16 \\
Good Carrots: 38 \\
Plan: \\
We can calculate the number of bad carrots by subtracting the number of good carrots from the total number of carrots picked. \\
Calculation: \\
Total Carrots Picked = Carol + Mom = 29 + 16 = 45 \\
Bad Carrots = Total Carrots Picked - Good Carrots = 45 - 38 = 7 \\
Answer: \\
There were 7 bad carrots.\\
\bottomrule
    \end{tabular}
\end{table*}
\endgroup
\begingroup
\begin{table*}[htb]
    \centering
    \caption{
    Example outputs by Zero-shot-PS+ for SVAMP. 
    }
    \vspace{2.8mm}
    \begin{tabular}{p{0.96\linewidth}}
        \toprule
        
\textbf{Q:} Mom buys 70 white t - shirts in total. If white t - shirts can be purchased in packages and mom buys 14 packages. How many white t - shirts does each package have?\\
\vspace{-1mm}
\textbf{A:} Let's first understand the problem, extract relevant variables and their corresponding numerals, and make and device a complete plan.Then, let's carry out the plan, calculate intermediate variables (pay attention to correct numerical calculation and commonsense), solve the problem step by step, and show the answer. \\
\vspace{-1mm}
\textbf{Output:}\\
Variables:\\
Total number of white t-shirts = 70\\
Number of packages = 14\\
Plan:\\
We can use the equation:\\
Total number of white t-shirts = Number of packages x Number of t-shirts per package\\
Calculation:\\
Number of t-shirts per package = 70/14 = 5\\
Answer:\\
Each package has 5 white t-shirts.\\
\midrule
\textbf{Q:} Jesse's rooms are 19 feet long and 18 feet wide. If she has 20 rooms in all. How much carpet does she need to cover the floors of all rooms?\\
\vspace{-1mm}
\textbf{A:} Let's first understand the problem, extract relevant variables and their corresponding numerals, and make and device a complete plan.Then, let's carry out the plan, calculate intermediate variables (pay attention to correct numerical calculation and commonsense), solve the problem step by step, and show the answer.\\
\vspace{-1mm}
\textbf{Output:}\\
Relevant Variables: \\
Number of rooms: 20\\
Length of each room: 19 feet\\
Width of each room: 18 feet\\
Plan:\\
Step 1: Calculate the area of each room.\\
Step 2: Multiply the area of each room by the number of rooms.\\
Step 3: Show the answer.\\
Calculation:\\
Step 1: Area of each room = Length x Width = 19 feet x 18 feet = 342 square feet\\
Step 2: Total area of all rooms = Area of each room x Number of rooms = 342 square feet x 20 rooms = 6840 square feet\\
Answer: Jesse needs 6840 square feet of carpet to cover the floors of all rooms.\\
\bottomrule
    \end{tabular}
\end{table*}
\endgroup
\begingroup
\begin{table*}[htb]
    \centering
    \caption{
    Example outputs by Zero-shot-PS+ for AddSub. 
    }
    \vspace{2.8mm}
    \begin{tabular}{p{0.96\linewidth}}
        \toprule
        
\textbf{Q:} There were 28 bales of hay in the barn. Tim stacked bales in the barn today. There are now 54 bales of hay in the barn. How many bales did he store in the barn ?\\
\vspace{-1mm}
\textbf{A:} Let's first understand the problem, extract relevant variables and their corresponding numerals, and devise a plan. Then, let's carry out the plan, calculate intermediate variables (pay attention to correct numeral calculation and commonsense), solve the problem step by step, and show the answer.
\\
\vspace{-1mm}
\textbf{Output:}\\
Given:\\
Initial number of bales of hay in the barn = 28\\
Number of bales of hay in the barn after Tim stacked them = 54\\
Plan:\\
Calculate the number of bales of hay Tim stored in the barn by subtracting the initial number of bales of hay from the number of bales of hay after Tim stacked them.\\
Calculation:\\
Number of bales of hay Tim stored in the barn = 54 - 28 = 26\\
Answer:\\
Tim stored 26 bales of hay in the barn.\\
\midrule
\textbf{Q:} There were 6 roses in the vase. Mary cut some roses from her flower garden. There are now 16 roses in the vase. How many roses did she cut ?\\
\vspace{-1mm}
\textbf{A:} Let's first understand the problem, extract relevant variables and their corresponding numerals, and devise a plan. Then, let's carry out the plan, calculate intermediate variables (pay attention to correct numeral calculation and commonsense), solve the problem step by step, and show the answer.\\
\vspace{-1mm}
\textbf{Output:}\\
Given:\\
Initial number of roses in the vase = 6\\
Final number of roses in the vase = 16\\
Plan:\\
Subtract the initial number of roses from the final number of roses to calculate the number of roses Mary cut.\\
Calculation:\\
Number of roses Mary cut = 16 - 6 = 10\\
Answer:\\
Mary cut 10 roses from her flower garden.\\
\bottomrule
    \end{tabular}
\end{table*}
\endgroup
\begingroup
\begin{table*}[htb]
    \centering
    \caption{
    Example outputs by Zero-shot-PS+ for SingleEq. 
    }
    \vspace{2.8mm}
    \begin{tabular}{p{0.96\linewidth}}
        \toprule
        
\textbf{Q:} After eating at the restaurant, Sally, Sam, and Alyssa decided to divide the bill evenly. If each person paid 45 dollars, what was the total of the bill ?\\
\vspace{-1mm}
\textbf{A:} Let's first understand the problem, extract relevant variables and their corresponding numerals, and devise a plan. Then, let's carry out the plan, calculate intermediate variables (pay attention to correct numeral calculation and commonsense), solve the problem step by step, and show the answer.
\\
\vspace{-1mm}
\textbf{Output:}\\
Problem: After eating at the restaurant, Sally, Sam, and Alyssa decided to divide the bill evenly. If each person paid 45 dollars, what was the total of the bill?\\
Variables:\\
- Total of the bill (T)\\
- Number of people (N)\\
- Amount paid by each person (A)\\
Numerals:\\
- N = 3\\
- A = 45\\
Plan:\\
- Calculate the total of the bill (T) by multiplying the number of people (N) by the amount paid by each person (A).\\
Calculation:\\
T = N x A\\
T = 3 x 45\\
T = 135\\
Answer: The total of the bill was 135 dollars.\\
\midrule
\textbf{Q:} Mary is baking a cake. The recipe wants 8 cups of flour. She already put in 2 cups. How many more cups does she need to add ?\\
\vspace{-1mm}
\textbf{A:} Let's first understand the problem, extract relevant variables and their corresponding numerals, and devise a plan. Then, let's carry out the plan, calculate intermediate variables (pay attention to correct numeral calculation and commonsense), solve the problem step by step, and show the answer.\\
\vspace{-1mm}
\textbf{Output:}\\
Mary needs to add 8 cups of flour to the cake. She has already put in 2 cups.\\
Plan:\\
Subtract the amount of flour Mary has already put in from the total amount of flour needed.\\
Calculation:\\
8 cups (total amount of flour needed) - 2 cups (amount of flour Mary has already put in) = 6 cups (amount of flour Mary needs to add)\\
Answer: Mary needs to add 6 more cups of flour.\\
\bottomrule
    \end{tabular}
\end{table*}
\endgroup
\begingroup
\begin{table*}[htb]
    \centering
    \caption{
    Example outputs by Zero-shot-PS+ for CommonsenseQA. 
    }
    \vspace{2.8mm}
    \begin{tabular}{p{0.96\linewidth}}
        \toprule
        
\textbf{Q:} What is a place that has a bench nestled in trees? Answer Choices: (A) state park (B) bus stop (C) bus depot (D) statue (E) train station\\
\vspace{-1mm}
\textbf{A:} Let's first understand the problem, extract relevant variables and their corresponding numerals, and devise a complete plan.Then, let's carry out the plan, calculate intermediate variables (pay attention to correct numerical calculation and commonsense), solve the problem step by step, and show the answer.\\
\vspace{-1mm}
\textbf{Output:}\\
The question is asking for a place that has a bench nestled in trees. The answer choices are (A) state park, (B) bus stop, (C) bus depot, (D) statue, and (E) train station.\\
The answer is (A) state park. State parks are typically located in natural areas with trees, and they often have benches for visitors to sit and enjoy the scenery.\\
\midrule
\textbf{Q:} Where is a good idea but not required to have a fire extinguisher? Answer Choices: (A) school bus (B) boat (C) house (D) hospital (E) school \\
\vspace{-1mm}
\textbf{A:} Let's first understand the problem, extract relevant variables and their corresponding numerals, and devise a complete plan.Then, let's carry out the plan, calculate intermediate variables (pay attention to correct numerical calculation and commonsense), solve the problem step by step, and show the answer.
\\
\vspace{-1mm}
\textbf{Output:}\\
The question is asking for a good idea to have a fire extinguisher, but it is not required. The answer choices are (A) school bus, (B) boat, (C) house, (D) hospital, and (E) school. \\
The best answer is (C) house. Having a fire extinguisher in a house is a good idea, but it is not required. Fire extinguishers are not typically required in school buses, boats, hospitals, or schools.
\\
\bottomrule
    \end{tabular}
\end{table*}
\endgroup
\begingroup
\begin{table*}[htb]
    \centering
    \caption{
    Example outputs by Zero-shot-PS+ for StrategyQA. 
    }
    \vspace{2.8mm}
    \begin{tabular}{p{0.96\linewidth}}
        \toprule
        
\textbf{Q:} Does the country that received the most gold medals during the 1976 Olympics still exist?\\
\vspace{-1mm}
\textbf{A:} Let's first prepare relevant information and make a plan. Then, let's answer the question step by step (pay attention to commonsense and logical coherence).\\
\vspace{-1mm}
\textbf{Output:}\\
Relevant information: \\
- The 1976 Olympics were held in Montreal, Canada. \\
- The country that received the most gold medals during the 1976 Olympics was the Soviet Union.\\
Plan: \\
1. Determine if the Soviet Union still exists. \\
2. If the Soviet Union still exists, answer the question. \\
3. If the Soviet Union does not exist, determine which countries it has been replaced by. Answer: \\
1. The Soviet Union no longer exists.\\ 
2. No, the country that received the most gold medals during the 1976 Olympics does not still exist. \\
3. The Soviet Union has been replaced by 15 independent countries, including Russia, Ukraine, Belarus, and Kazakhstan.\\
\midrule
\textbf{Q:}Is the Mona Lisa in the same museum as the Venus de Milo? \\
\vspace{-1mm}
\textbf{A:} Let's first prepare relevant information and make a plan. Then, let's answer the question step by step (pay attention to commonsense and logical coherence).\\
\vspace{-1mm}
\textbf{Output:}\\
Relevant information: \\
Mona Lisa: Located in the Louvre Museum in Paris, France\\
Venus de Milo: Located in the Louvre Museum in Paris, France\\
Plan: \\
Step 1: Identify the locations of the Mona Lisa and the Venus de Milo\\
Step 2: Compare the locations\\
Answer: \\
Yes, the Mona Lisa and the Venus de Milo are both located in the Louvre Museum in Paris, France.\\
\bottomrule
    \end{tabular}
\end{table*}
\endgroup
\begingroup
\begin{table*}[htb]
    \centering
    \caption{
    Example outputs by Zero-shot-PS+ for Last Letters. 
    }
    \vspace{2.8mm}
    \begin{tabular}{p{0.96\linewidth}}
        \toprule
        
\textbf{Q:} Take the last letters of each words in "Lino Mariel Aditya Elisabeth" and concatenate them\\
\vspace{-1mm}
\textbf{A:}Let's devise a plan and solve the problem step by step. \\
\vspace{-1mm}
\textbf{Output:}\\
Step 1: Take the last letters of each word in the given string.\\

Lino: o\\
Mariel: l\\
Aditya: a\\
Elisabeth: h\\

Step 2: Concatenate the last letters.\\

Answer: olah\\
\midrule
\textbf{Q:} Take the last letters of each words in "Jacky Socorro Mark Wanda" and concatenate them\\
\vspace{-1mm}
\textbf{A:}Let's devise a plan and solve the problem step by step.\\
\vspace{-1mm}
\textbf{Output:}\\
Step 1: Take the last letters of each word.\\

Jacky: y\\
Socorro: o\\
Mark: k\\
Wanda: a\\

Step 2: Concatenate the letters.\\

Answer: yoka
\\
\bottomrule
    \end{tabular}
\end{table*}
\endgroup
\begingroup
\begin{table*}[htb]
    \centering
    \caption{
    Example outputs by Zero-shot-PS+ for Coin Flip. 
    }
    \vspace{2.8mm}
    \begin{tabular}{p{0.96\linewidth}}
        \toprule
        
\textbf{Q:} A coin is heads up. Walter does not flip the coin. Mikayla does not flip the coin. Larry does not flip the coin. Ryan does not flip the coin. Is the coin still heads up? Note that "flip" here means "reverse".
\\
\vspace{-1mm}
\textbf{A:} Let's first understand the problem and devise a complete plan. Then, let's carry out the plan and reason problem step by step. Every step answer the subquestion, "does the person flip and what is the coin's current state?". According to the coin's last state, give the final answer (pay attention to every flip and the coin’s turning state).
\\
\vspace{-1mm}
\textbf{Output:}\\
Plan:\\
Step 1: Walter does not flip the coin.\\
Step 2: Mikayla does not flip the coin.\\
Step 3: Larry does not flip the coin.\\
Step 4: Ryan does not flip the coin.\\
Answer: Yes, the coin is still heads up.\\
Explanation:\\
Step 1: Walter does not flip the coin. Therefore, the coin is still heads up.\\
Step 2: Mikayla does not flip the coin. Therefore, the coin is still heads up.\\
Step 3: Larry does not flip the coin. Therefore, the coin is still heads up.\\
Step 4: Ryan does not flip the coin. Therefore, the coin is still heads up.\\
Conclusion: Yes, the coin is still heads up.\\
\midrule
\textbf{Q:} A coin is heads up. Lorena does not flip the coin. Shana does not flip the coin. Priscilla does not flip the coin. Summer does not flip the coin. Is the coin still heads up? Note that "flip" here means "reverse"
 \\
\vspace{-1mm}
\textbf{A:} Let's first understand the problem and devise a complete plan. Then, let's carry out the plan and reason problem step by step. Every step answer the subquestion, "does the person flip and what is the coin's current state?". According to the coin's last state, give the final answer (pay attention to every flip and the coin’s turning state).\\
\vspace{-1mm}
\textbf{Output:}\\
Plan:\\
Step 1: Lorena does not flip the coin.\\
Step 2: Shana does not flip the coin.\\
Step 3: Priscilla does not flip the coin.\\
Step 4: Summer does not flip the coin.\\

Answer:\\
Step 1: Lorena does not flip the coin. Therefore, the coin is still heads up.\\
Step 2: Shana does not flip the coin. Therefore, the coin is still heads up.\\
Step 3: Priscilla does not flip the coin. Therefore, the coin is still heads up.\\
Step 4: Summer does not flip the coin. Therefore, the coin is still heads up.\\

Final Answer: Yes, the coin is still heads up.\\
\bottomrule
    \end{tabular}
\end{table*}
\endgroup

\end{document}